\documentclass{article}
\usepackage{spconf,amsmath,epsfig}
\usepackage{xcolor}
\usepackage{amssymb}
\usepackage{multirow}
\usepackage{leading}
\usepackage{hyperref}
\usepackage{svg}

\let\OLDthebibliography\thebibliography
\renewcommand\thebibliography[1]{
  \OLDthebibliography{#1}
  \setlength{\parskip}{0pt}
  \setlength{\itemsep}{0pt plus 0.3ex}
}

\begin{document}\sloppy

\def\x{{\mathbf x}}
\def\L{{\cal L}}

\title{Unsupervised Domain Adaptation Learning for Hierarchical \\Infant Pose Recognition with Synthetic Data}
%
\name{Cheng-Yen Yang$^1$, Zhongyu Jiang$^1$, Shih-Yu Gu$^1$, Jenq-Neng Hwang$^1$, Jang-Hee Yoo$^2$}
\address{$1$ Department of Electrical and Computer Engineering, University of Washington, USA;\\
$2$ Electronics and Telecommunications Research Institute, South Korea}

\maketitle

\begin{abstract}

The Alberta Infant Motor Scale (AIMS) is a well-known assessment scheme that evaluates the gross motor development of infants by recording the number of specific poses achieved. With the aid of the image-based pose recognition model, the AIMS evaluation procedure can be shortened and automated, providing early diagnosis or indicator of potential developmental disorder. Due to limited public infant-related datasets, many works use the SMIL-based method to generate synthetic infant images for training. However, this domain mismatch between real and synthetic training samples often leads to performance degradation during inference. In this paper, we present a CNN-based model which takes any infant image as input and predicts the coarse and fine-level pose labels. The model consists of an image branch and a pose branch, which respectively generates the coarse-level logits facilitated by the unsupervised domain adaptation and the 3D keypoints using the HRNet with SMPLify optimization. Then the outputs of these branches will be sent into the hierarchical pose recognition module to estimate the fine-level pose labels. We also collect and label a new AIMS dataset, which contains 750 real and 4000 synthetic infants images with AIMS pose labels. Our experimental results show that the proposed method can significantly align the distribution of synthetic and real-world datasets, thus achieving accurate performance on fine-grained infant pose recognition.


\end{abstract}
\section{Introduction}
\label{sec:intro}

Autism Spectrum Disorder (ASD) is a developmental disability that can cause significant social, communication, and behavioral challenges. Recent medical research suggests that early signs of ASDs may first manifest within the motor control system and present as a motor delay. According to the studies \cite{ASD1, ASD2, ASD3}, infants who have delays in acquiring motor skills are at higher risk of developing ASD and may serve as an early indicator of neuro-developmental disorder. To better measure or quantify the level of motor development , Alberta Infants Motor Scale (AIMS) \cite{AIMS} is introduced as a scale assessment procedure to evaluate and track infant motor milestones based on counting the number of gross motor skills that the target can achieve. However, traditional AIMS assessment requires trained professionals to conduct, which are considered time-consuming and inefficient. Moreover, the collection of enough and diversified infant pose images for algorithmic development is challenging due to privacy concerns and institutional Review Board (IRB) regulations. To overcome these issues, a systematic AIMS assessment system, which incorporates the deep learning based 3D infant pose estimation trained by synthetic infant pose data and unsupervised domain adaptation technologies, is proposed to show promising performance for effective AIMS assessment of real-world infant in this paper. 

\begin{figure}[t]
\centering
\centerline{\epsfig{figure=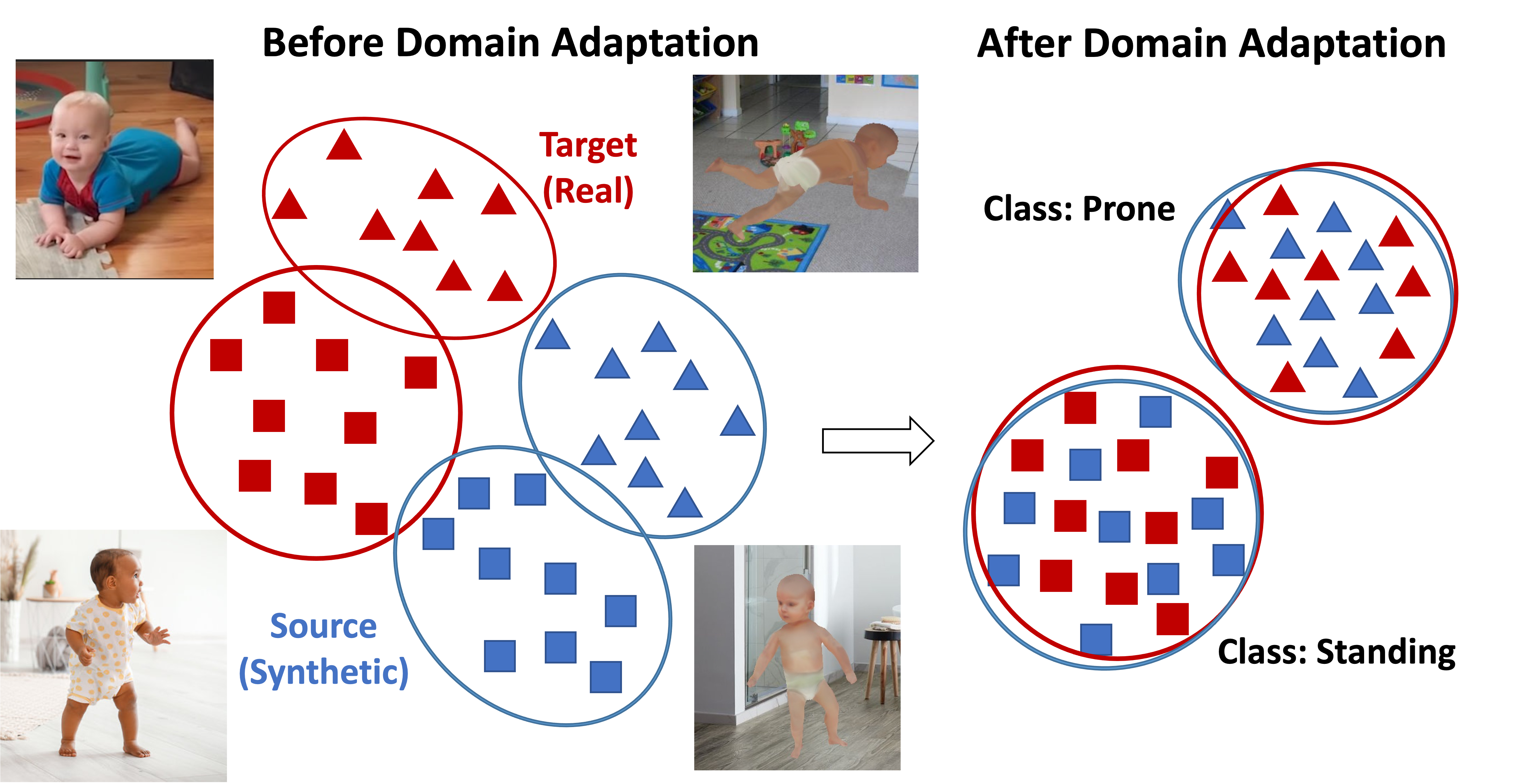,width=\linewidth}}
\label{fig:res}
\vspace{-1em}
\caption{Unsupervised domain adaptation can exploit the local affinity to capture the fine-grained information and align the distribution accordingly, which can significantly boost the performance of systematic AIMS assessment of real infant images using synthetic infant training data.}
\vspace{-2em}
\end{figure}

\let\thefootnote\relax\footnotetext{* This work was supported by Institute of Information \& Communications Technology Planning \& Evaluation (IITP) grant funded by the Korea government (MSIT) (2019-0-00330). }


Due to the difficulty in collecting and annotating such fine-grained poses, many datasets collect long untrimmed sequences that lack several distinct poses. Moreover, real-world infant-related data are extremely limited because of privacy concerns and institutional Review Board (IRB) regulations, resulting in most related research on infants using the synthetic datasets for model training. However, one of the problems of training using synthetic data is the domain shift between the synthetic data and real-world data, which often leads to performance degradation and poor generalization ability of the trained model. One attempt to solve such a problem is transferring a model learned on a labeled source domain to an unlabeled target domain, known as Unsupervised Domain Adaptation (UDA). Specifically, we have synthetic infant images with pose labels as the source domain and real-world infant images without labels as the target domain. By using UDA, our final goal is to align the source and target domains in the feature space for further applications.

Our contributions are in two-folds: (1) Present a new infant pose dataset with both synthetic and real-world images with fine-level annotation labels, that allow us to evaluate the cross-dataset generalization ability of the model, and (2) Integrate an unsupervised domain adaptation (UDA) algorithm into the hierarchical pose recognition framework to enable transfer learning across domains for better feature extractions of the existing CNN framework.

The paper is organized in the following: We review some related works to our research in Sec. \ref{sec:related}. Then we describe our AIMS infant pose dataset synthesizing process in Section \ref{sec:dataset}. In Sections \ref{sec:methods} and \ref{sec:exp}, the methods and experimental results of the infant pose recognition will be presented, followed by the conclusion in Section \ref{sec:conclusion}.

\section{Related Works}
\label{sec:related}

\subsection{Human Pose Estimation \& Recognition}

\noindent \textbf{Pose Estimation} In general, 2D human pose estimation methods can be classiﬁed into bottom-up and top-down approaches. Top-down approaches \cite{conv_pose_machine, stacked_hourglass, hrnet} first detect human bounding boxes and then perform human keypoint detection within every bounding box's region. On the other hand, bottom-up approaches \cite{openpose, personlab, HigherHRNet} first detect all keypoints on all humans in the image and then assign keypoints belonging to the same person of their owner. 


Most recent 3D human pose estimation works take 2D skeletons as input.  Bogo et al.\cite{simplify} adopt the Skinned Multi-Person Linear (SMPL) \cite{SMPL}, a statistical body shape  model, as an initial human skeleton and optimize the skeleton by minimizing the reprojection error to get the final 3D human pose. Zhao et al.\cite{gcn3dpose} adopt a Graph Convolutional Network (GCN) to process the features of 2D joints as node features to generate 3D joint predictions. Pavllo et al.\cite{videopose3d} propose the most well-known temporal-based 3D solutions, the VideoPose3D, which adopt 2D joints of hundreds of frames as input to predict a single 3D skeleton. We take advantage of Skinned Multi-Infant Linear (SMIL)\cite{SMIL}, which is an infant version of SMPL, to generate our synthetic dataset, resulting in an image-based infant pose recognition task. SMPLify\cite{simplify}, which minimizes the error between the 3D model joints and detected 2D joints, is adopted for our 3D infant pose estimation.

\noindent \textbf{Pose Recognition} Vision-based human pose recognition aims to obtain posture and predict the corresponding action from input images or video sequences. Most pose or action recognition works are developed closely with pose estimation by leveraging the skeleton-based approaches for prediction \cite{ar_1, ar_2}. Current publicly available human pose or action datasets are predominantly from scenes such as sports and daily activities performed by adult humans \cite{MPII, LSP} which differ dramatically from the infant motions in terms of achievable poses. 

\subsection{Unsupervised Domain Adaptation}


Various works have been targeting domain adaptation to overcome the domain shift problems. Sener et al. \cite{UDA_Sener} propose to use clustering techniques and pseudo-labels to obtain discriminative features. Taigman et al. \cite{UDA_Taigman} propose cross-domain image translation methods. Ganin et al. \cite{UDA_Ghifary} propose a representative method of distribution matching involving training a domain classifier using the intermediate features and generating the features that deceive the domain classifier. This method utilizes the similar techniques used in generative adversarial networks (GANs). The category classifier is trained to predict the task-specific category labels. And the domain classifier is trained to predict the domain of each input. The two classifiers share feature extraction layers that are trained to predict the label of source samples correctly and deceive the domain classifier. Thus, the distributions of the intermediate  features of the target and source samples are made similar. However, an issue of unsupervised domain adaptation by back-propagation is that the target features can be near a task-specific classifier’s boundary, which will cause the target samples far from source ones (ambiguous features) to be likely misclassified after alignment.

\subsection{Infant Dataset}

Privacy remains one of the issues for infant-related dataset collection process. Therefore, the majority of the existing infant datasets are synthetic images. Currently, there are only limited infant-related datasets: MINI-RGBD \cite{MINI-RGBD}, SyRIP \cite{FiDIP}, and Zhou et al. \cite{HIPC}. MINI-RGBD mapped real infant movements to the SMIL model, generating RGB and depth video sequences with 2D and 3D joint coordinates. However, these data are synthesized from infants under seven months old and thus present simple poses with small changes over samples. SyRIP is composed of two portions, real and synthetic. The real part consists of 700 infant images collected from public sources like Youtube and Google, while the synthetic part consists of 1000 images rendered using the SMIL model. The 2D joint coordinates are fully annotated in COCO format \cite{COCO} for these images, which is a huge contribution toward the infant pose estimation research but lack labels for the pose classification task. Zhou's dataset contains 5500 synthetic images with 11 classes selected from AIMS, but the insufficient corresponding real-world evaluation portion makes it difficult to justify the model's generalization ability.
\section{AIMS Dataset}
\label{sec:dataset}

To provide an evaluation of the generalization ability of the models, we need a dataset that includes pose labels of the real-world infant image. We start from a small number of real-world infant samples and use the SMIL-based model to enlarge our real-world infant dataset with labeled synthetic data. 


\subsection{Skinned Multi-Infant Linear (SMIL) Model}

Skinned Multi-Person Linear Body (SMPL) \cite{SMPL} is a skinned vertex-based model that is able to represent different human body shapes and poses with the parameters learned from data. Skinned Multi-Infant Linear body model (SMIL) \cite{SMIL} is a derived version of the SMPL learned from the sequences of freely moving infants in \cite{MINI-RGBD}. Some pose coefficients $\theta \in \mathbb{R}^{3\times N_j}$ and the shape coefficients $\beta \in \mathbb{R}^{N_s}$  serve as input, and the output is a mesh consisting of $N_v=6890$ vertices with $N_j=24$ joints defined in SMPL. 

In order to obtain realistic pose and shape parameters for infants, we take advantage of SMPLify \cite{SMPLify} to better fit any given infant image by minimizing the overall loss function:

\begin{equation}
    L(\theta, \beta) = L_{J_{2D}} + L_{\theta} + L_{\beta},
\end{equation}

\noindent where $L_{J_{2D}}$ denotes the distance between estimated 2D joints and the 2D projection of the 3D joints. $L_{\theta}$ and $L_{\beta}$ respectively denotes the  simple $L_2$ prior for body pose and body shape.

\subsection{Rendering}

\begin{table}[t]
\centering
\caption{The selected infant poses in AIMS \cite{AIMS} in our work with 4 coarse-level and 12 fine-level labels.}
\vspace{.3em}
\begin{tabular}{|c|l|}
\hline
Coarse-Level              & \multicolumn{1}{c|}{Fine-Level} \\ \hline
\multirow{4}{*}{Prone}    & Prone Lying                     \\ \cline{2-2} 
                          & Forearm Support                 \\ \cline{2-2} 
                          & Reciprocal Crawling             \\ \cline{2-2} 
                          & Four-Point Kneeing              \\ \hline
\multirow{3}{*}{Supine}   & Supine Lying                    \\ \cline{2-2} 
                          & Hands to Knee/Feet              \\ \cline{2-2} 
                          & Rolling                         \\ \hline
\multirow{3}{*}{Sitting}  & Sitting w/ Support              \\ \cline{2-2} 
                          & Sitting w/ Arm Support          \\ \cline{2-2} 
                          & Sitting w/o Support             \\ \hline
\multirow{2}{*}{Standing} & Four-Point Standing             \\ \cline{2-2} 
                          & Standing                        \\ \hline
\end{tabular}
\end{table}

After fitting SMIL model for each infant instance in the image, we can then render synthetic images using different texture of the infant model, different backgrounds, and reasonable translation operations. The ground-truth 3D keypoint coordinates $\textbf{X}_{3d}$ can be obtained directly from the fitted SMIL model, where all 3D keypoints will be normalized with respect to the distance from nose to pelvis to ensure the scaling consistency. 

\begin{figure}[h]

  \centering
 \centerline{\epsfig{figure=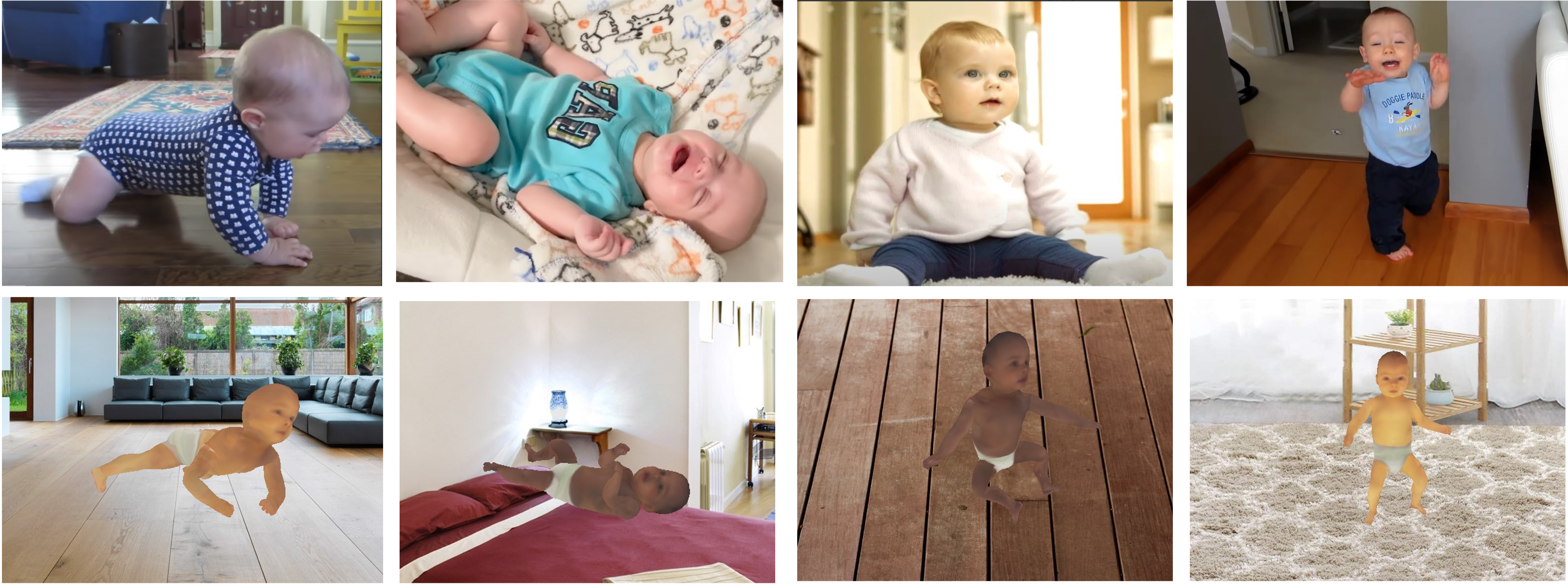,width=0.9\linewidth}}
\caption{Sample images from our collected AIMS dataset. The first row is  from the real portion and the second row is the synthetic portion.}
\end{figure}

As for 2D keypoint coordinates, we need to reproject the 3D keypoints in the world coordinates to the image coordinates based on  the camera parameters  used to render each synthetic sample. The ground-truth 2D keypoint coordinates $\textbf{x}_{2d}$ can thus be obtained by:

\begin{equation}
    \textbf{x}_{2d} = \textbf{K} \cdot [\textbf{R}|\textbf{T}] \cdot \textbf{X}_{3d},
\end{equation}

\noindent where $\textbf{K}$ and $[\textbf{R}|\textbf{T}]$ are the pre-defined camera intrinsic and extrinsic parameters.

\begin{figure*}[!htb]
\centering
\includegraphics[width=1.05\textwidth]{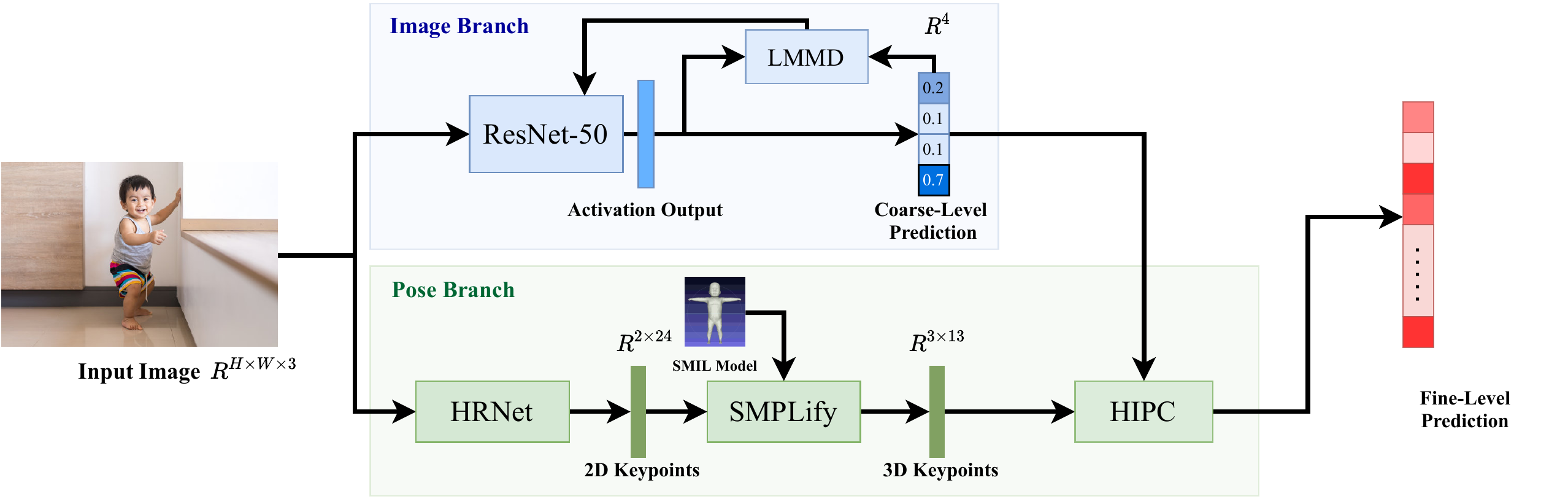}
\vspace{-1em}
\caption{The overview of pipeline of our proposed infant action recognition with unsupervised domain adaptation learning.}
\label{fig:2}
\end{figure*}


We extend the real portion of the SyRIP dataset \cite{FiDIP} by annotating and categorizing the real infant portion into 12 selected fine-level gross motor poses, a very small portion ($\approx 5\%$) of samples are withdrawn due to the poses not falling into any of defined fine-level poses. We randomly assign different camera parameters and remove those unnatural samples after syntheses. In addition, we also collect 200 background images from Google, and select 1000 scenes from INDOOR dataset \cite{indoor_dataset} under the labels like bedroom, children room, and nursery, to mimic the real-world data and prevent overfitting on the synthetic images.

In total, the entire synthetic portion consists of around 4000 samples, while the real portion consists of 750 samples with ground-truth 2D keypoints (image coordinate), 3D keypoints (world coordinate), coarse-level and fine-level AIMS labels.

\section{Methodology}
\label{sec:methods}

\subsection{Image Branch: Domain Adaptation Network}

For image-level classification, the gap between real-world infant samples and synthetic infant samples often causes performance degradation and leads to inaccurate predictions. We can formulate the infant pose recognition task as  an unsupervised domain adaptation scenario. More specifically, given a source domain $D_s=\{ (\textbf{x}^s, \textbf{y}^s)  \}$ and an unlabeled target domain $D_t=\{ \textbf{x}^t  \}$, which are sampled from different distributions $s$ and $t$, respectively. The goal of unsupervised domain adaption (UDA) is to train the CNN classifier that reduces the discrepancy of the two distributions. Specifically, we adopt  Local Maximum Mean Discrepancy (LMMD) to aid our UDA on aligning the relevant subdomain distributions of domain specific layer activation across different domains.

The unbiased estimator of LMMD can be expressed as:

\begin{equation}
\label{eq:LMMD}
    d_{\mathcal{H}} (s,t) =  \frac{1}{n_c} \sum_{c \in C} \Big|\Big| \sum_{x^s \in \mathcal{D}_s} w^{s}_{c} \phi (x^s) - \sum_{x^t \in \mathcal{D}_t} w^{t}_{c} \phi (x^t)\Big|\Big|^2_{\mathcal{H}},
\end{equation}  

\noindent where $\mathcal{H}$ is the reproducing kernel Hillbert space (RKHS) \cite{RKHS} with kernel $k$. The kernel $l$ represents$k(x^s, x^t)=\langle\phi(x^s), \phi(x^t)\rangle$, which is the inner product of two vectors of some feature mapping operation $\phi(\cdot)$. 

The important characteristic of LMMD is the $w^{s}_{c}$ and $w^{t}_{c}$, which denote the weights of $x$ belonging to a given class $c$ in the source and target domains, respectively. The weights for source domain $w^{s}_{c}$ $w^{s}_{c}$can be easily computed using the  ground truth label as an one-hot vector $w^{s}_{c}=y^{sc}_{i}/\sum_{i\neq j} y^{sc}_{j}$. However for the unlabeled target domain, the weights $w^{t}_{c}$ have to be adaptively estimated using the output $z$ of each activation layer $l \in L$ in order to compute Eq. \ref{eq:LMMD}  .

Finally, the adaptation loss will be multiplied by a coefficient $\lambda$ and added to the classification loss, which is the naive cross-entropy loss. The overall loss  to be minimized  becomes:

\begin{equation}
    \mathcal{L}_{overall} = \mathcal{L}_{classify} + \lambda \cdot \mathcal{L}_{adapt}.
\end{equation}

\begin{figure}[t]

 \centering
 \centerline{\epsfig{figure=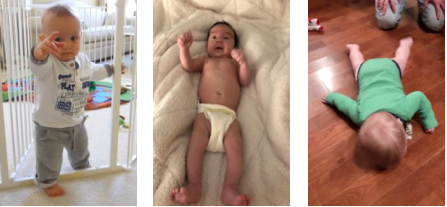,width=.65\linewidth}}
 \vspace{0.2cm}
 \centerline{\epsfig{figure=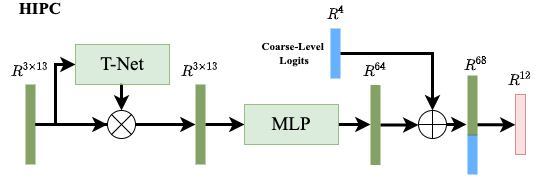,width=.9\linewidth}}
\label{fig:confusion}
\vspace{-1em}
\caption{An example of different poses that result in the same 2D keypoints in different view angle and the detailed architecture of HIPC aiming to alleviate such problem by using the logits from the image branch.}
\end{figure}

\begin{table}[h]
\caption{Experimental results of our proposed method for coarse-level and fine-level classification. The \textbf{bold} text denotes the highest top-1 accuracy achieved under the same experiment setting.}
\small
\label{table:main}
\begin{tabular}{c|lc}
\hline
Task & \multicolumn{1}{c}{Method} & Top-1 Accurcay  \\ \hline
\multirow{3}{*}{\begin{tabular}[c]{@{}c@{}}Coarse-Level\\ Classification\end{tabular}} & Image Branch & 75.5\% \\
     & Image Branch (w/ LMMD)     & \textbf{85.3\%} \\
     & Pose Branch                & 83.3\%          \\ \hline
\multirow{4}{*}{\begin{tabular}[c]{@{}c@{}}Fine-Level\\ Classification\end{tabular}}   & Image Branch & 40.4\% \\
     & Image Branch (w/ LMMD)     & 50.5\%          \\
     & Pose Branch                & 68.0\%          \\
     & \textbf{Ours}                       & \textbf{76.8\%} \\ \hline
\end{tabular}
\end{table}

\subsection{Pose Branch: 2D/3D Pose Estimation}
We adopt HRNet\cite{hrnet} and SMPLify\cite{simplify} for our 2D and 3D pose estimation. For 2D Pose Estimation, HRNet takes a single image $I$ as input, and generates heatmaps $H$ for all the keypoints.  The coordinates $\textbf{x}_{2d}$ can be obtained by finding points with the highest values in $H$. After getting $\textbf{x}_{2d}$, SMPLify\cite{simplify} is used to optimize the SMIL model iteratively by minimizing the reprojection error of projected $\textbf{X}_{3d}$ with $\textbf{x}_{2d}$.

\subsection{HIPC: Hierarchical Infant Pose Classifier}

For infant pose recognition, to overcome the confusion caused by the viewing perspectives, 3D skeletons should be used for better recognition performance. However, as Fig \ref{fig:confusion}, unlike adult human pose recognition, the view angles of infant images are more flexible, and two similar 3D skeletons may lead to two completely different fine level poses. As a result, we take advantage of a Hierarchical Infant Pose Classifier (HIPC)\cite{HIPC}, which takes the 3D keypoints $\textbf{X}_{3d}$ and coarse-level recognition logits as input, for getting better fine level recognition result.

\section{Experimental Results}
\label{sec:exp}




To evaluate the performance of both coarse-level and fine-level infant pose recognition tasks, we train the model using all of the synthetic dataset along with  a subset of the ground truth labelled real dataset, where 4 coarse labels and 12 fine sub-labels with a total of 4000 samples are used. We evaluate our model using the test subset of the real data with a total of 198 images and record the Top-1 accuracies. The codes are implemented in Pytorch and we conduct the experiments on one Nvidia GeForce GTX Titan XP card.

\begin{table}[t]
\centering
\small
\label{table:lambda}
\caption{Effect of domain adaptation loss when using different training dataset (i.e., different source domain distribution).}
\begin{tabular}{c|lc}
\hline
\begin{tabular}[c]{@{}c@{}}Source\\ Domain\end{tabular} & \multicolumn{1}{c}{Method} & \begin{tabular}[c]{@{}c@{}}Top-1 Accuracy\\ (Coarse-Level)\end{tabular} \\ \hline
\multirow{2}{*}{\textit{Synthetic}}      & Image Branch           & $45.3\%$ \\
                                & Image Branch (w/ LMMD) & $\textbf{60.3\%}$ \\ \hline
\multirow{2}{*}{\textit{Synthetic+Real}} & Image Branch           & $75.5\%$ \\
                                & Image Branch (w/ LMMD) & $\textbf{85.3\%}$ \\ \hline
\end{tabular}
\end{table}

\begin{figure}[t]

\centering
 \centerline{\epsfig{figure=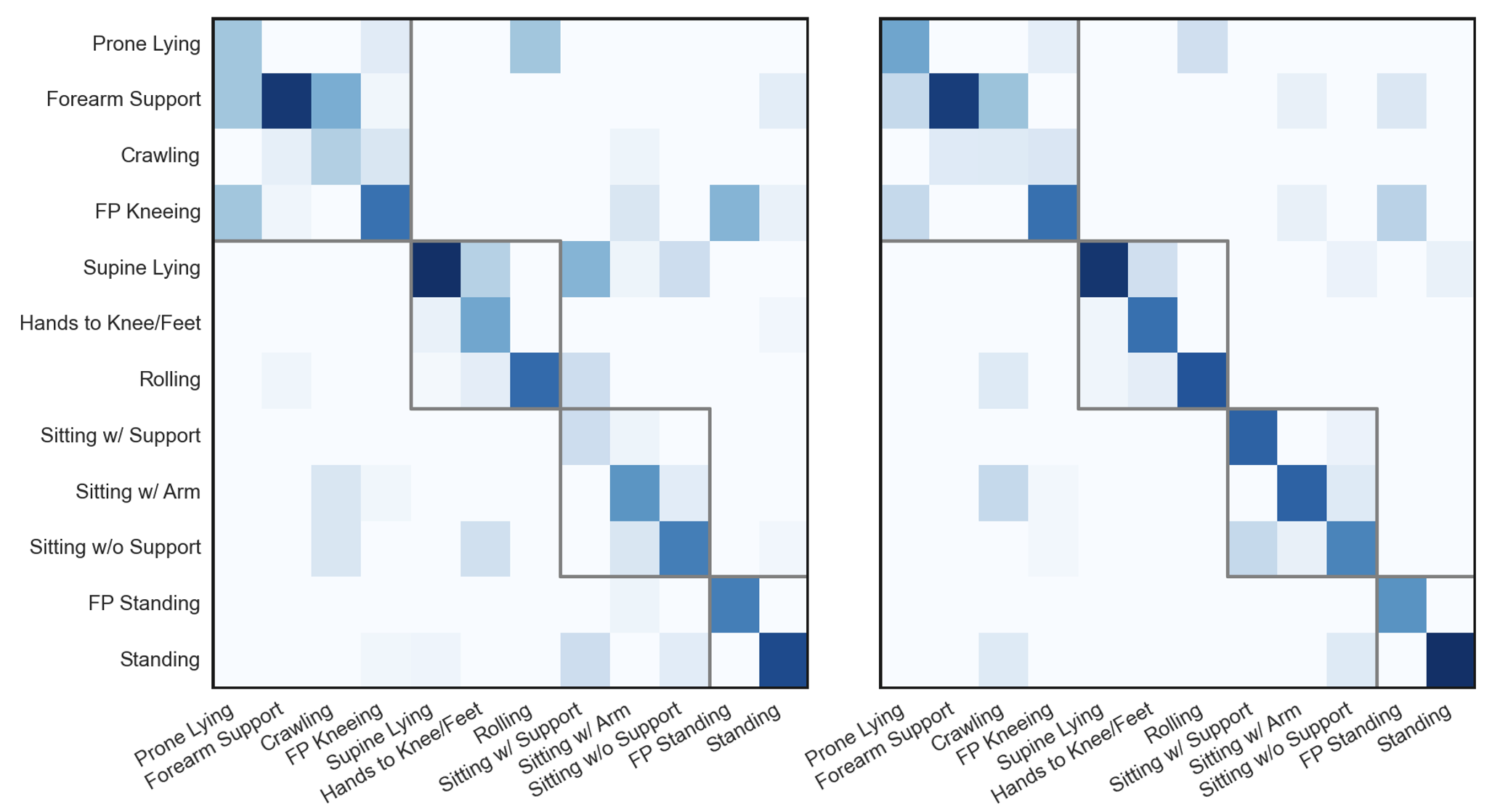,width=1.0\linewidth}}
\label{fig:confusion}
\vspace{-1em}
\caption{A comparison of the confusion matrices for fine-level classification results from baseline (left) and our purposed method (right). The gray boxes represent the fine-level labels with the same coarse-level label.}
\end{figure}

\subsection{Performance of Infant Pose Recognition}

 The experimental results are shown in Table \ref{table:main}, which shows the top-1 accuracies for coarse-level pose classification (4 classes) and fine-level pose classification (12 classes) based on the same algorithmic configurations. With the help of adding LMMD loss and  a $\lambda$ of $0.5$, the domain adaptation module introduced to the naive ResNet-50 is able to improve the performance of our image-branch coarse-level classification from $75.5\%$ to $85.3\%$ (using \textit{Syn+Real}). Moreover, the fine-level classification results can be improved from $68.0\%$ to $76.8\%$ after using the logits from the image-branch to guide the  prediction from pose branch. In Fig. \ref{fig:confusion}, we can clearly see that most of those misclassified samples across coarse-levels (i.e., outside of the gray bounding boxes) had been corrected with the aid of logits from image-branch in our proposed method.

Our best model leverages both unsupervised domain adaptation for coarse-level classification and hierarchical pose recognition framework and can truly overcome the source-target domain distribution mismatch, as we achieved a $76.8\%$ accuracy on fine-level classification. 

\subsection{Ablation Study}

We now analyze the effect of domain adaptation on different training dataset. The source domain denotes the dataset used during training is denoted as \textit{Synthetic}, which represents the entire synthetic portion from AIMS dataset and \textit{Synthetic+Real} represents the setting of adding a small number of real-world infant samples into the source domain. As shown in Table \ref{table:lambda}, LMMD loss can significantly improve the performance of classification without any major modification being made to the CNN-based backbone, since the adaptation loss is  computed using the activation outputs and pseudo-labels only (i.e., from $45.3\%$ to $60.3\%$ Top-1 accuracies on \textit{Synthetic} and from $75.5\%$ to $85.3\%$ Top-1 accuracies on \textit{Synthetic+Real}).


\section{Conclusion}
\label{sec:conclusion}


Fine-level infant pose recognition may help doctors or parents determine infants' motor skill development. In this paper, we proposed a new AIMS Dataset for fine-level infant pose recognition, with both synthetic and real-world data. With this dataset, we integrated an unsupervised domain adaptation algorithm into the hierarchical pose recognition framework to enable transfer learning across domains for better feature extractions of existing CNN framework, and finally, we achieved $76.8\%$ Top-1 accuracy with the domain-adopted model on the AIMS test dataset.



\section{References}
\small
\bibliographystyle{IEEEbib}
\bibliography{icme2022template}

\begin{thebibliography}{10}

\bibitem{ASD1}
Anjana~N. Bhat, Rebecca~J. Landa, and James C.~(Cole) Galloway,
\newblock ``{Current Perspectives on Motor Functioning in Infants, Children,
  and Adults With Autism Spectrum Disorders},''
\newblock {\em Physical Therapy}, vol. 91, no. 7, pp. 1116--1129, 07 2011.

\bibitem{ASD2}
Joanne Flanagan, Rebecca Landa, Anjana Bhat, and Margaret Bauman,
\newblock ``Head lag in infants at risk for autism: A preliminary study,''
\newblock {\em The American journal of occupational therapy}, vol. 66, pp.
  577--85, 09 2012.

\bibitem{ASD3}
James Patterson, Vickie Armstrong, Eric Duku, Annie Richard, Martina Franchini,
  Jessica Brian, Lonnie Zwaigenbaum, Susan Bryson, Lori-Ann Sacrey, Caroline
  Roncadin, and Isabel Smith,
\newblock ``Early trajectories of motor skills in infant siblings of children
  with autism spectrum disorder,''
\newblock {\em Autism Research}, 11 2021.

\bibitem{AIMS}
Piper Martha and Darrah Johanna,
\newblock {\em Motor Assessment of the Developing Infant},
\newblock Saunders, 1994.

\bibitem{conv_pose_machine}
Shih-En Wei, Varun Ramakrishna, Takeo Kanade, and Yaser Sheikh,
\newblock ``Convolutional pose machines,''
\newblock in {\em Proceedings of the IEEE conference on Computer Vision and
  Pattern Recognition}, 2016, pp. 4724--4732.

\bibitem{stacked_hourglass}
Alejandro Newell, Kaiyu Yang, and Jia Deng,
\newblock ``Stacked hourglass networks for human pose estimation,''
\newblock in {\em European conference on computer vision}. Springer, 2016, pp.
  483--499.

\bibitem{hrnet}
Jingdong Wang, Ke~Sun, Tianheng Cheng, Borui Jiang, Chaorui Deng, Yang Zhao,
  Dong Liu, Yadong Mu, Mingkui Tan, Xinggang Wang, et~al.,
\newblock ``Deep high-resolution representation learning for visual
  recognition,''
\newblock {\em IEEE transactions on pattern analysis and machine intelligence},
  2020.

\bibitem{openpose}
Zhe Cao, Gines Hidalgo, Tomas Simon, Shih-En Wei, and Yaser Sheikh,
\newblock ``Openpose: realtime multi-person 2d pose estimation using part
  affinity fields,''
\newblock {\em IEEE TPAMI}, vol. 43, no. 1, pp. 172--186, 2019.

\bibitem{personlab}
George Papandreou, Tyler Zhu, Liang-Chieh Chen, Spyros Gidaris, Jonathan
  Tompson, and Kevin Murphy,
\newblock ``Personlab: Person pose estimation and instance segmentation with a
  bottom-up, part-based, geometric embedding model,''
\newblock in {\em ECCV}, 2018, pp. 269--286.

\bibitem{HigherHRNet}
Bowen Cheng, Bin Xiao, Jingdong Wang, Honghui Shi, Thomas~S. Huang, and Lei
  Zhang,
\newblock ``Bottom-up higher-resolution networks for multi-person pose
  estimation,''
\newblock {\em CoRR}, vol. abs/1908.10357, 2019.

\bibitem{simplify}
Federica Bogo, Angjoo Kanazawa, Christoph Lassner, Peter Gehler, Javier Romero,
  and Michael~J Black,
\newblock ``Keep it smpl: Automatic estimation of 3d human pose and shape from
  a single image,''
\newblock in {\em ECCV}, 2016, pp. 561--578.

\bibitem{SMPL}
Matthew Loper, Naureen Mahmood, Javier Romero, Gerard Pons-Moll, and Michael~J.
  Black,
\newblock ``{SMPL}: A skinned multi-person linear model,''
\newblock {\em ACM Trans. Graphics (Proc. SIGGRAPH Asia)}, vol. 34, no. 6, pp.
  248:1--248:16, Oct. 2015.

\bibitem{gcn3dpose}
Long Zhao, Xi~Peng, Yu~Tian, Mubbasir Kapadia, and Dimitris~N. Metaxas,
\newblock ``Semantic graph convolutional networks for 3d human pose
  regression,''
\newblock in {\em Proceedings of the IEEE/CVF CVPR}, June 2019.

\bibitem{videopose3d}
Dario Pavllo, Christoph Feichtenhofer, David Grangier, and Michael Auli,
\newblock ``3d human pose estimation in video with temporal convolutions and
  semi-supervised training,''
\newblock in {\em Proceedings of the IEEE/CVF CVPR}, June 2019.

\bibitem{SMIL}
Nikolas Hesse, Sergi Pujades, Javier Romero, Michael~J. Black, Christoph
  Bodensteiner, Michael Arens, Ulrich~G. Hofmann, Uta Tacke, Mijna
  Hadders-Algra, Raphael Weinberger, Wolfgang M\"uller-Felber, and A.~Sebastian
  Schroeder,
\newblock ``Learning an infant body model from {RGB-D} data for accurate full
  body motion analysis,''
\newblock in {\em International Conference on Medical Image Computing and
  Computer-Assisted Intervention (MICCAI)}. Springer, 2018.

\bibitem{ar_1}
Limin Wang, Yuanjun Xiong, Zhe Wang, Yu~Qiao, Dahua Lin, Xiaoou Tang, and Luc
  {Val Gool},
\newblock ``Temporal segment networks: Towards good practices for deep action
  recognition,''
\newblock in {\em ECCV}, 2016.

\bibitem{ar_2}
Haodong Duan, Yue Zhao, Kai Chen, Dian Shao, Dahua Lin, and Bo~Dai,
\newblock ``Revisiting skeleton-based action recognition,''
\newblock {\em CoRR}, vol. abs/2104.13586, 2021.

\bibitem{MPII}
Mykhaylo Andriluka, Leonid Pishchulin, Peter Gehler, and Bernt Schiele,
\newblock ``2d human pose estimation: New benchmark and state of the art
  analysis,''
\newblock in {\em IEEE Conference on Computer Vision and Pattern Recognition
  (CVPR)}, June 2014.

\bibitem{LSP}
Sam Johnson and Mark Everingham,
\newblock ``Clustered pose and nonlinear appearance models for human pose
  estimation,''
\newblock in {\em Proceedings of the British Machine Vision Conference}, 2010,
\newblock doi:10.5244/C.24.12.

\bibitem{UDA_Sener}
Ozan Sener, Hyun~Oh Song, Ashutosh Saxena, and Silvio Savarese,
\newblock ``Learning transferrable representations for unsupervised domain
  adaptation,''
\newblock in {\em NIPS}, 2016.

\bibitem{UDA_Taigman}
Yaniv Taigman, Adam Polyak, and Lior Wolf,
\newblock ``Unsupervised cross-domain image generation,''
\newblock {\em ArXiv}, vol. abs/1611.02200, 2017.

\bibitem{UDA_Ghifary}
Muhammad Ghifary, W.~Bastiaan Kleijn, Mengjie Zhang, David Balduzzi, and Wen
  Li,
\newblock ``Deep reconstruction-classification networks for unsupervised domain
  adaptation,''
\newblock {\em CoRR}, vol. abs/1607.03516, 2016.

\bibitem{MINI-RGBD}
Nikolas Hesse, Christoph Bodensteiner, Michael Arens, Ulrich~G. Hofmann,
  Raphael Weinberger, and A.~Sebastian Schroeder,
\newblock ``Computer vision for medical infant motion analysis: State of the
  art and {RGB-D} data set,''
\newblock in {\em ECCV 2018 Workshops}. 2018, Springer International
  Publishing.

\bibitem{FiDIP}
Xiaofei Huang, Nihang Fu, Shuangjun Liu, and Sarah Ostadabbas,
\newblock ``Invariant representation learning for infant pose estimation with
  small data,''
\newblock in {\em IEEE International Conference on Automatic Face and Gesture
  Recognition (FG), 2021}, December 2021.

\bibitem{HIPC}
Jianxiong Zhou, Zhongyu Jiang, Jang-Hee Yoo, and Jenq-Neng Hwang,
\newblock ``Hierarchical pose classification for infant action analysis and
  mental development assessment,''
\newblock in {\em ICASSP 2021 - 2021 IEEE International Conference on
  Acoustics, Speech and Signal Processing (ICASSP)}, 2021, pp. 1340--1344.

\bibitem{COCO}
Tsung{-}Yi Lin, Michael Maire, Serge~J. Belongie, Lubomir~D. Bourdev, Ross~B.
  Girshick, James Hays, Pietro Perona, Deva Ramanan, Piotr Doll{\'{a}}r, and
  C.~Lawrence Zitnick,
\newblock ``Microsoft {COCO:} common objects in context,''
\newblock {\em CoRR}, vol. abs/1405.0312, 2014.

\bibitem{SMPLify}
Federica Bogo, Angjoo Kanazawa, Christoph Lassner, Peter Gehler, Javier Romero,
  and Michael~J. Black,
\newblock ``Keep it {SMPL}: Automatic estimation of {3D} human pose and shape
  from a single image,''
\newblock in {\em ECCV 2016}. Oct. 2016, Lecture Notes in Computer Science,
  Springer International Publishing.

\bibitem{indoor_dataset}
Ariadna Quattoni and Antonio Torralba,
\newblock ``Recognizing indoor scenes,''
\newblock in {\em 2009 IEEE Conference on Computer Vision and Pattern
  Recognition}, 2009, pp. 413--420.

\bibitem{RKHS}
S.K. Zhou and R.~Chellappa,
\newblock ``From sample similarity to ensemble similarity: probabilistic
  distance measures in reproducing kernel hilbert space,''
\newblock {\em IEEE Transactions on Pattern Analysis and Machine Intelligence},
  vol. 28, no. 6, pp. 917--929, 2006.

\end{thebibliography}

\end{document}